\documentclass[runningheads]{llncs}
\usepackage{amsmath,amssymb,amsfonts}
\usepackage{graphicx}
\usepackage{notoccite}
\usepackage{hyperref}
\usepackage{cleveref,multirow}
\usepackage[dvipsnames]{xcolor}
\usepackage{adjustbox}

\usepackage{silence}
\begin{document}
\title{CompTLL-UNet: Compressed Domain Text-Line Localization in Challenging Handwritten Documents using Deep Feature Learning from JPEG Coefficients}
\author{Bulla Rajesh\inst{1,2},\orcidID{0000-0002-5731-9755} \and
Sk Mahafuz Zaman\inst{2} \and
Mohammed Javed\inst{2}\orcidID{0000-0002-3019-7401} \and P. Nagabhushan\inst{2}\orcidID{0000-0002-4638-5482}}

\authorrunning{B. Rajesh et al.}
\institute{
Department of CSE, IIIT-SriCity, Chittoor, AP, India, \and Department of IT, IIIT-Allahabad, Prayagraj, UP, India\\
\email{rajesh.bulla@iiits.in,\{mit2020005, javed, pnagabhushan\}@iiita.ac.in}}
\maketitle             
\begin{abstract}

Automatic localization of text-lines in handwritten documents is still an open and challenging research problem. Various writing issues such as uneven spacing between the lines, oscillating and touching text, and the presence of skew become much more challenging when the case of complex handwritten document images are considered for segmentation directly in their respective compressed representation. This is because, the conventional way of processing compressed documents is through decompression, but here in this paper, we propose an idea that employs deep feature learning directly from the JPEG compressed coefficients without full decompression to accomplish text-line localization in the JPEG compressed domain. A modified U-Net architecture known as Compressed Text-Line Localization Network (CompTLL-UNet) is designed to accomplish it. The model is trained and tested with JPEG compressed version of benchmark datasets including ICDAR2017 (cBAD) and ICDAR2019 (cBAD), reporting the state-of-the-art performance with reduced storage and computational costs in the JPEG compressed domain. 

\end{abstract}

\section{Introduction}  
\begin{figure}[!ht]
    \centering
    \includegraphics[scale=.50]{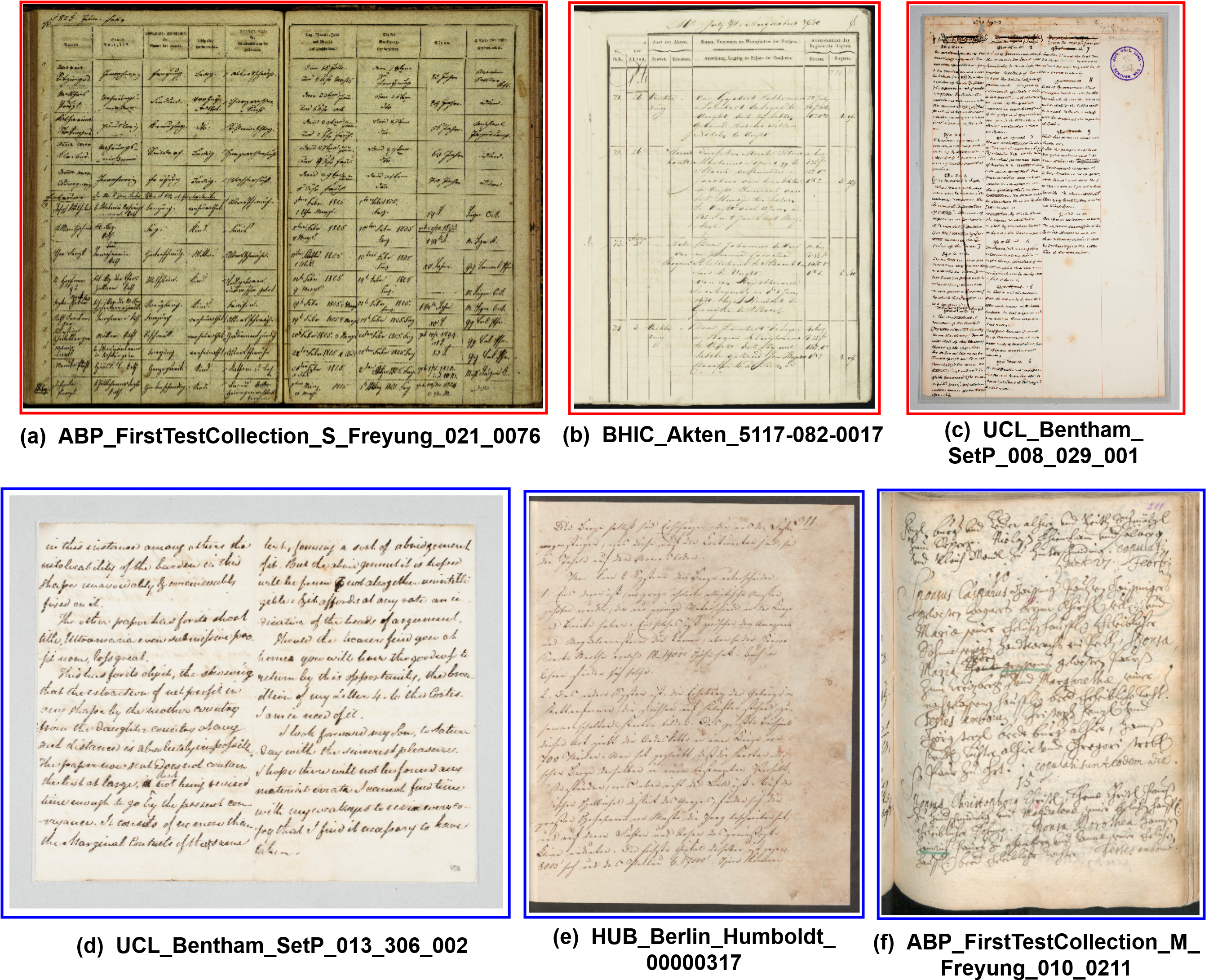}
    \caption{Sample complex handwritten document images reproduced from ICDAR2017 dataset}
    \label{figone}
\end{figure}

Localization of text-lines in the document image is a very crucial pre-processing step towards many significant Document Image Analysis (DIA) applications like word spotting, handwriting recognition, and Optical Character Recognition (OCR) frequently required in many public places like Banks, Postal service and embassies etc. \cite{javedicdarword,Tejasvee}. Specifically, in the case of handwritten documents, the text-lines usually have skew and uneven spacing, and due to which many a time the characters get touching and overlapping, thus making segmentation a challenging problem \cite{renton2018fully,text-linefully,bulla2020}. Apart from these inherent issues, sometimes the contents in the historical handwritten document come up with colossal noise, degraded text, variations of backgrounds, complex layouts, multi-columns, presence of tables, and marginalia, as shown in Figure \ref{figone}. Performing segmentation or localization task on such documents further increase the complexity of many challenges \cite{8978147}. However, the processing of documents directly in compressed form is an important research issue because compressing images before transmission or archival has become the normal trend to save disk space, transmission time and internet bandwidth \cite{javed2018review,mukhopadhyay2011image,bisen2023segmentation}. Therefore, researchers have started looking for technology that can handle or process these compressed images without involving decompression stage that requires more computational overhead \cite{gueguen2018faster,ehrlich2019deep,javedicdarword,rajesh2019dct}. In the recent literature \cite{liu2022semantic,bisen2023segmentation}, there are some efforts to process document images directly in the compressed domain and reported less computation time and reduced storage space. Therefore performing text line localization in such complex handwritten documents and that too directly in the compressed domain is going to be advantageous and contributing research problem for all DIA applications. Although analyzing compressed representation is a difficult task, a typical algorithm that is proposed should be able to correctly locate text-lines irrespective of various challenges present in it. Also, the direct processing of compressed data should show a improved performance in terms of reduced computational and storage costs. 

\begin{figure*}[!ht]
    \centering
    \includegraphics[scale=0.60]{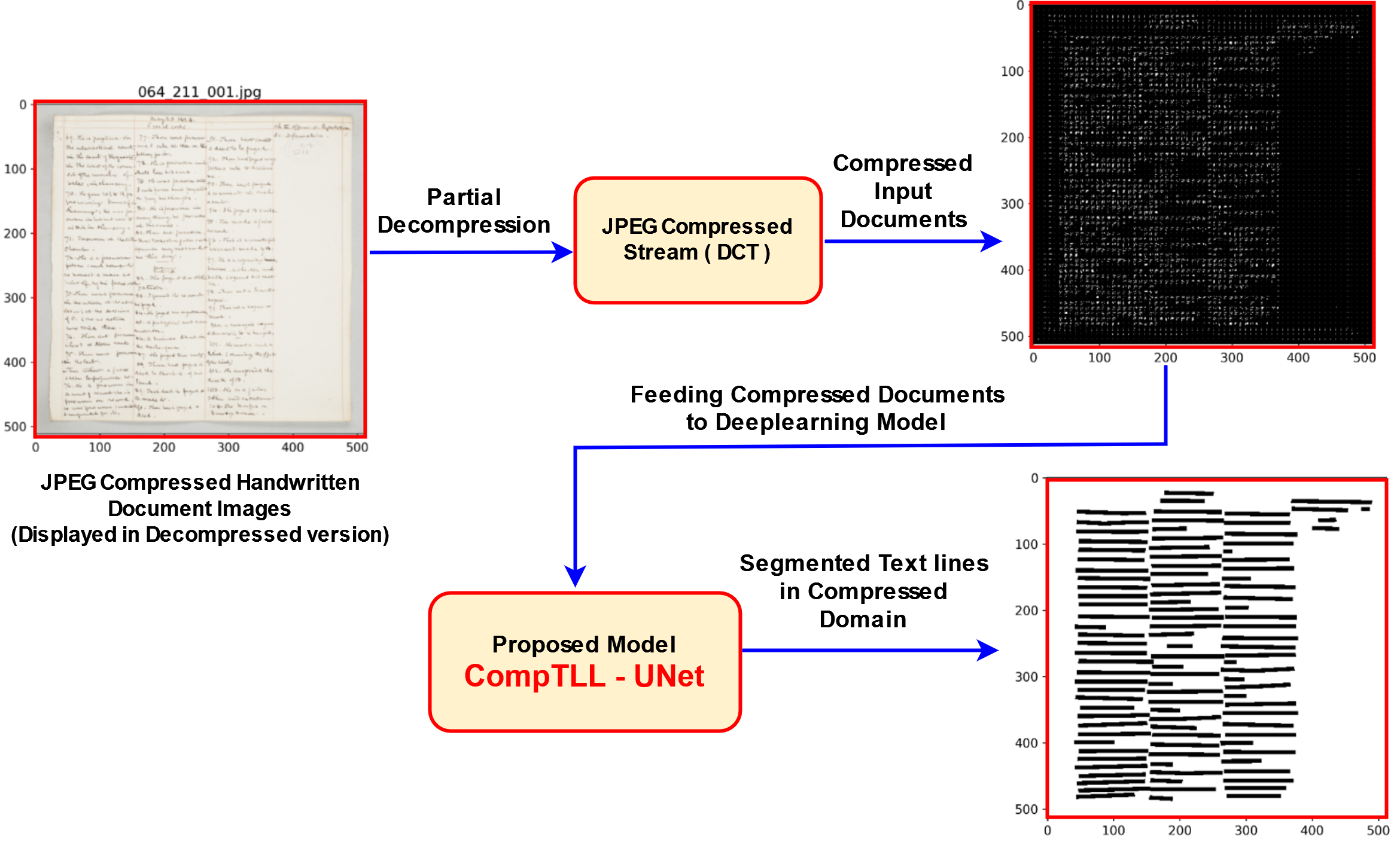}
    \caption{The flow diagram of the proposed model CompTTL-UNet for localizing the text-lines directly in JPEG compressed document images.}
    \label{figtwo}
\end{figure*}
Therefore, this research paper proposes to tackle the issue of text-line localization and explore text-line segmentation in the case of complex historical handwritten documents directly in the JPEG compressed representation. Unlike the conventional approaches, the proposed method in the present paper applies a partial decompression to extract the JPEG compressed streams of document images, and feeds the input stream directly to the deep learning architectures as shown in Figure \ref{figtwo}. 

There are three major contributions reported in this research paper:
\begin{itemize}
    \item CompTLL-UNet model for text-line localization in compressed domain.
    \item Direct feeding of JPEG compressed stream into the CompTTL-UNet model for deep feature learning from compressed data. The mathematical visualization of the processing (in the form of matrix and kernals) of compressed streams and pixel streams has been explained.
    \item Reduced computational and storage costs in comparison to conventional methods.
\end{itemize}
The proposed model has been trained and tested on two benchmark datasets, and the results have showed the state-of-the-art accuracy and significant performance in terms of computational cost and storage as explained in detailed in the result section. The rest of the paper is divided into four sections. Section 2 discusses the extraction of JPEG Compressed input stream and feeding to the deep learning model. Section 3 explains the proposed methodology and details of model architecture. Section 4 report the experimental results and analysis, and comparison with existing methods. Section 5 concludes the work with a brief summary and possible future work.

\section{Preamble to JPEG compressed domain}
\hspace{20pt}This section demonstrates the differences between feeding of uncompressed document images and JPEG compressed document images into the deep learning model, and contrasts the challenges associated with JPEG compressed document images against uncompressed document images.
\subsection{Uncompressed Document Images}
\begin{figure*}[!ht]
    \centering
    \includegraphics[scale=0.20]{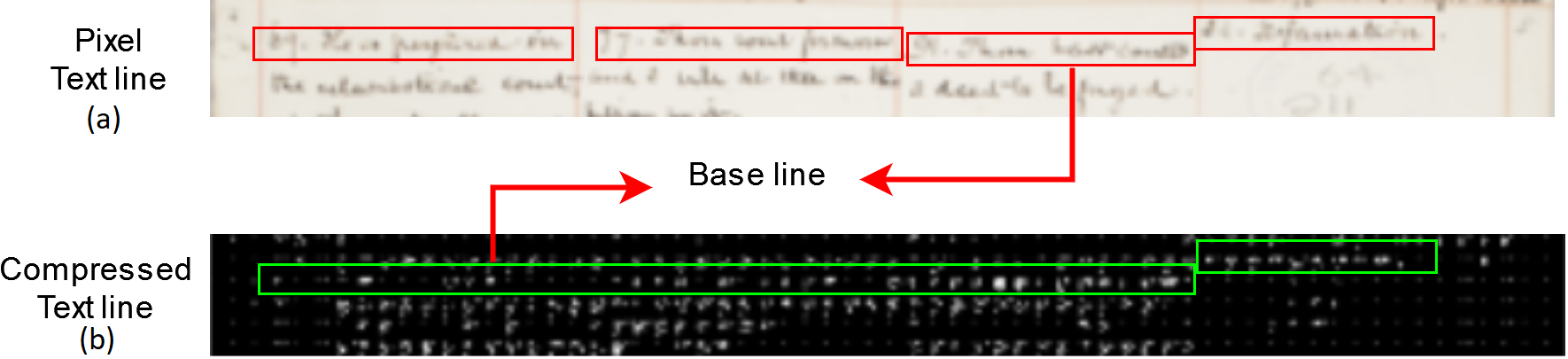}
    \caption{The visualization of base line regions of text-lines in (a) pixel/uncompressed document and (b) in JPEG compressed document images (064\_211\_001.jpg).}
    \label{figurethree}
\end{figure*}

\hspace{20pt}So far, many research problems including feature extraction, segmentation and recognition have been discussed extensively in uncompressed domain and addressed many implementation challenges associated with each problem in the case of uncompressed document images\cite{liu2022semantic,10008841}. But the challenges with compressed documents are different in comparison to pixel/uncompressed domain, complex to visualize and much more difficult to address \cite{liu2022semantic}. Before looking into the JPEG compressed document images, some common observation available on the uncompressed document images provides background knowledge to understand the JPEG compressed input streams as shown in Figure \ref{figurethree}. The figure 3(a) shows the baseline regions of a sample text-lines in an uncompressed document image (064\_211\_001.jpg) and figure 3(b) shows the JPEG compressed representation of figure 3(a). 

 In the figure 3(a), it can be noticed that the text-lines are arranged in four columns with irregular separating spaces and touching of the text with adjacent text-lines. Similarly, in figure 3(b) same text-lines which are in the compressed representation not providing any clues of separating space. In order to understand the pixel level details of the same text-lines, a sample pixel values from an $8\times8$ block of the a text-line is shown in $B$ matrix. 
 \begin{sloppypar}
\begin{equation}
B =
\begin{bmatrix} 
244 & 244 & 244 & 244 & 244 & 244 &  \\
244 & 244 & 244 & 244 & 244 & 244 &  \\
244 & 244 & 244 & 244 & 244 & 244 &  \\
244 & 244 & 244 & 244 & 244 & 244 &  \\
244 & 244 & 244 & 244 & 244 & 244 &  \\
244 & 244 & 244 & 244 & 244 & 244 &  
\end{bmatrix}
K = \frac{1}{9}
\begin{bmatrix} 
1 & 1 & 1 \\
1 & 1 & 1 \\
1 & 1 & 1 
\end{bmatrix}
\end{equation}
\end{sloppypar}
 \hspace{20pt} In $B$, it is observed that the pixel values possess high correlation, positive integers and less chaos/variation. Since deep neural networks are well known for processing the regular pixel values with different filters $K$ of sizes $3\times3$, $5\times5$, and $7\times7$, and the networks steadily pool the meaningful features when filters are applied on these pixels at each level in the network. It continue to maintain the pixels correlation it finds as the network goes deep, and eventually learns salient feature representations at every level. However, this process may looks similar in the compressed domain, but that is not the same in the case of JPEG compressed document images because the arrangement of text-line in the compressed representation are resulted from $8\times8$ DCT transformation without bothering about the positions of the inside text contents, which is explained below.


\subsection{JPEG Compressed Document Images}
\hspace{20pt}During compression, JPEG algorithm divides the uncompressed image $f$ into $8\times8$ blocks and transforms each block $B$ using Discrete Cosine Transformation (DCT) as per Eq (\ref{eqone}) and quantization \cite{gueguen2018faster}, and generates $DB$ (DCT block) and $QDB$ (Quantized DCT block) representations. And further it applies sequence of operations such as DPCM, run-length, and Huffman entropy encodings to compress the contents to its binary form. Since, JPEG compressed images undergo DCT transformation, quantization and subsequent encodings, most of the text contents get overlapped during $8\times8$ block division and loose some useful visual clues.  Some times the text-line get mixed up with adjacent (all sides) text-lines without leaving any separation gap for baseline regions as shown in Figure \ref{figurethree}(b) (green color bounding box). In comparison to the text-line contents in uncompressed documents (a) (red color bounding box), most text-line contents in compressed documents appears in a crooked way as shown for a sample text-line in JPEG compressed domain in Figure \ref{figurethree}. Here, the text-lines in different columns are mixed, and only few minute visual clues are available to analyze. In comparison to pixel block $B$, the $DB$ and $QDB$ blocks contain very few coefficient values and when a kernal $K$ is applied when fed to deep learning model it process same set of values as shown in $DB\textsuperscript{*}$ and $QDB\textsuperscript{*}$ blocks. However it is observed that since these coefficient values are already showing some average spatial behaviour of the block, deep models are optimizing based on the those available values in the DCT stream.    

\begin{equation}
 F_{uv} = \frac{c_{u}c_{v}}{4} \sum_{i=0}^{7} \sum_{j=0}^{7} B(i,j) cos(\frac{(2i+1)u\pi}{16}) cos(\frac{(2j+1)v\pi}{16})
\label{eqone} 
\end{equation}
$$
{ \scriptstyle
Where,\hspace{10mm} 
C_{u}, C_{v} =
\begin{cases}
\frac{1}{\sqrt 2}, & for \hspace{5pt} u, v=0 \\
1, & otherwise,
\end{cases}
}
$$


\begin{small}
\begin{equation}
    DB = 
\begin{bmatrix} 
1955 & 0 & 0 & 0 & 0 & 0 & 0 & 0 &  \\
0 & 0 & 0 & 0 & 0 & 0 & 0 & 0 &  \\
-1 & 0 & 0 & 0 & 0 & 0 & 0 & 0 &  \\
0 & 0 & 0 & 0 & 0 & 0 & 0 & 0 &  \\
0 & 0 & 0 & 0 & 0 & 0 & 0 & 0 &  \\
0 & 0 & 0 & 0 & 0 & 0 & 0 & 0 &  \\
0 & 0 & 0 & 0 & 0 & 1 & 0 & 0 &  \\
0 & 0 & 0 & 0 & 0 & 0 & 0 & 0 &  \\
\end{bmatrix}
DB\textsuperscript{*} =
\begin{bmatrix} 
217 & 0 & 0 & 0 & 0 & 0 &  \\
0 & 0 & 0 & 0 & 0 & 0 &  \\
0 & 0 & 0 & 0 & 0 & 0 &  \\
0 & 0 & 0 & 0 & 0 & 0 &  \\
0 & 0 & 0 & 0 & 0 & 0 &  \\
0 & 0 & 0 & 0 & 0 & 0 &  \\
\end{bmatrix}
\end{equation}
\end{small}


\begin{equation}
QDB = 
\begin{bmatrix} 
122 & 0 & 0 & 0 & 0 & 0 & 0 & 0 &  \\
0 & 0 & 0 & 0 & 0 & 0 & 0 & 0 &  \\
0 & 0 & 0 & 0 & 0 & 0 & 0 & 0 &  \\
0 & 0 & 0 & 0 & 0 & 0 & 0 & 0 &  \\
0 & 0 & 0 & 0 & 0 & 0 & 0 & 0 &  \\
0 & 0 & 0 & 0 & 0 & 0 & 0 & 0 &  \\
0 & 0 & 0 & 0 & 0 & 0 & 0 & 0 &  \\
0 & 0 & 0 & 0 & 0 & 0 & 0 & 0 &  \\
\end{bmatrix}
QDB\textsuperscript{*} =
\begin{bmatrix} 
13 & 0 & 0 & 0 & 0 & 0 &  \\
0 & 0 & 0 & 0 & 0 & 0 &  \\
0 & 0 & 0 & 0 & 0 & 0 &  \\
0 & 0 & 0 & 0 & 0 & 0 &  \\
0 & 0 & 0 & 0 & 0 & 0 &  \\
0 & 0 & 0 & 0 & 0 & 0 &  \\
\end{bmatrix}
\end{equation}

\section{Proposed Methodology}
\hspace{20pt}This section explains the proposed methodology and the details of the deep learning architecture to localize the text-line boundaries in challenging handwritten documents directly in JPEG compressed domain.

\hspace{20pt}The sequence of steps in the proposed method are shown in Figure \ref{figtwo}. First the partial decompression (entropy decoding) is applied on JPEG compressed document images to extract the input JPEG compressed streams. This input stream is arranged in quantized representation ($QDB$) in which except DC most of the AC coefficient values are zero. Then the extracted streams are fed as input to a proposed deep learning model CompTLL-UNet to learn different patterns to localize the baseline regions of text-line boundaries in the JPEG compressed document image.  

\hspace{20pt}Since UNet \cite{ronneberger2015u} is a popular architecture for segmentation tasks, the proposed architecture in the present paper is designed by modifying the existing UNet \cite{ronneberger2015u} architecture. The architecture of UNet is modified in such a way that it can be trained with compressed data and still produce promising accuracy. In the UNet architecture, the successive pooling layers are replaced by up-sampling layers so the successive convolution layer can learn more precise information. It has a large number of feature channels that propagate context information to higher resolution layers. Since the input in the present paper is extracted from compressed streams, the architecture should be redesigned to learn the features from the compressed streams. 
Based on the experimental study with different parameter settings, the first layer of the UNet model has been modified with 64 channels and employed a stride of 7$\times$7 to observe the DCT patterns at the 1st layer. Average pooling has been applied for down-sampling as most of the DCT coefficients are zero, and less context for preserving the context for successive layers. The average pooling technique has learned more details from the compressed streams and propagates them to the successive consecutive layers.

The model architecture contains 19 convolution layers and 4 layers of convolution transpose. After each layer of convolution, the input stream is normalized with batch normalization and spatial dropout. For every 2 layers of convolution, we have used Max-Pool, ReLU activation, for down sampling the input, and convolution transpose and concatenation are used for up sampling. In the last layer the Sigmoid is used as an activation function. Furthermore, the output image has the same resolution as the input image. In our case, it's $512 \times 512 $ with one channel.


\section{Experiments and Analysis}
\hspace{20pt}The proposed model has been experimented on two benchmark datasets ICDAR2017 (cBAD) and ICDAR2019. The document images in ICDAR2017 cBAD dataset contains two types of layouts: simple layouts [Track A] and complex layouts [Track B]. The basis of cBAD contains document images from 9 different archives as shown some of them in Figure \ref{figone}. There were 216 images in TrackA and 267 images in TrackB with proper annotations. The second dataset is ICDAR2019, which is the extended version of ICDAR2017. In this there are total 1510 images with additional images and more number of additional challenges.

\hspace{20pt}The model has been trained and tested on JPEG compressed version of the ICDAR2017 and ICDAR2019 datasets individually. The datasets are split into two parts where first part contains 90\% of images used for training and second part contains 10\% of images used for testing.  The proposed model has been designed using Keras framework, and trained on Google Colab pro platform with NVIDIA T4 GPU. The compressed input streams are resized into $512 \times 512$ to feed to deep learning model CompTLL-UNet. The model has been trained and tested on both the datasets.The model has been trained for 50 epochs with a batch size of 5. The performance of proposed model is evaluated by three types of standard and popular \cite{renton2018fully} metrics Precision, Recall and F-Measure given in Eq (\ref{Preci}), Eq (\ref{Recal}) and Eq (\ref{fmes}). 

\begin{multicols}{2}
  \begin{equation}
    Precision = \frac{TP}{TP+FP} \times 100 
\label{Preci}
  \end{equation}
  \begin{equation}
   Recall = \frac{TP}{TP+FN} \times 100 
    \label{Recal}
  \end{equation}
\end{multicols}

\begin{equation}
F-Measure = 2 \times \frac{Precision \times Recall}{Precision+Recall} \times 100\\
\label{fmes}
\end{equation}
Since the proposed deep learning model is a segmentation problem we have also evaluated the robustness of the proposed model based on DICE score and IoU as given in Eq (\ref{dice}) and Eq (\ref{iou}). Where DICE score is the area overlapped between ground truth and predicted image divided by the number of pixels in two images. And IoU is known as "Intersection over Union", and it specifies an overlapping between ground truth and predicted result.
\begin{multicols}{2}
  \begin{equation}
  DICE = \frac{2TP}{2TP+FP+FN}
\label{dice}
  \end{equation}
  \begin{equation}
IoU = \frac{TP}{TP+FP+FN}
\label{iou}
  \end{equation}
\end{multicols}

\begin{figure*}[!ht]
    \centering
    \includegraphics[scale=.34]{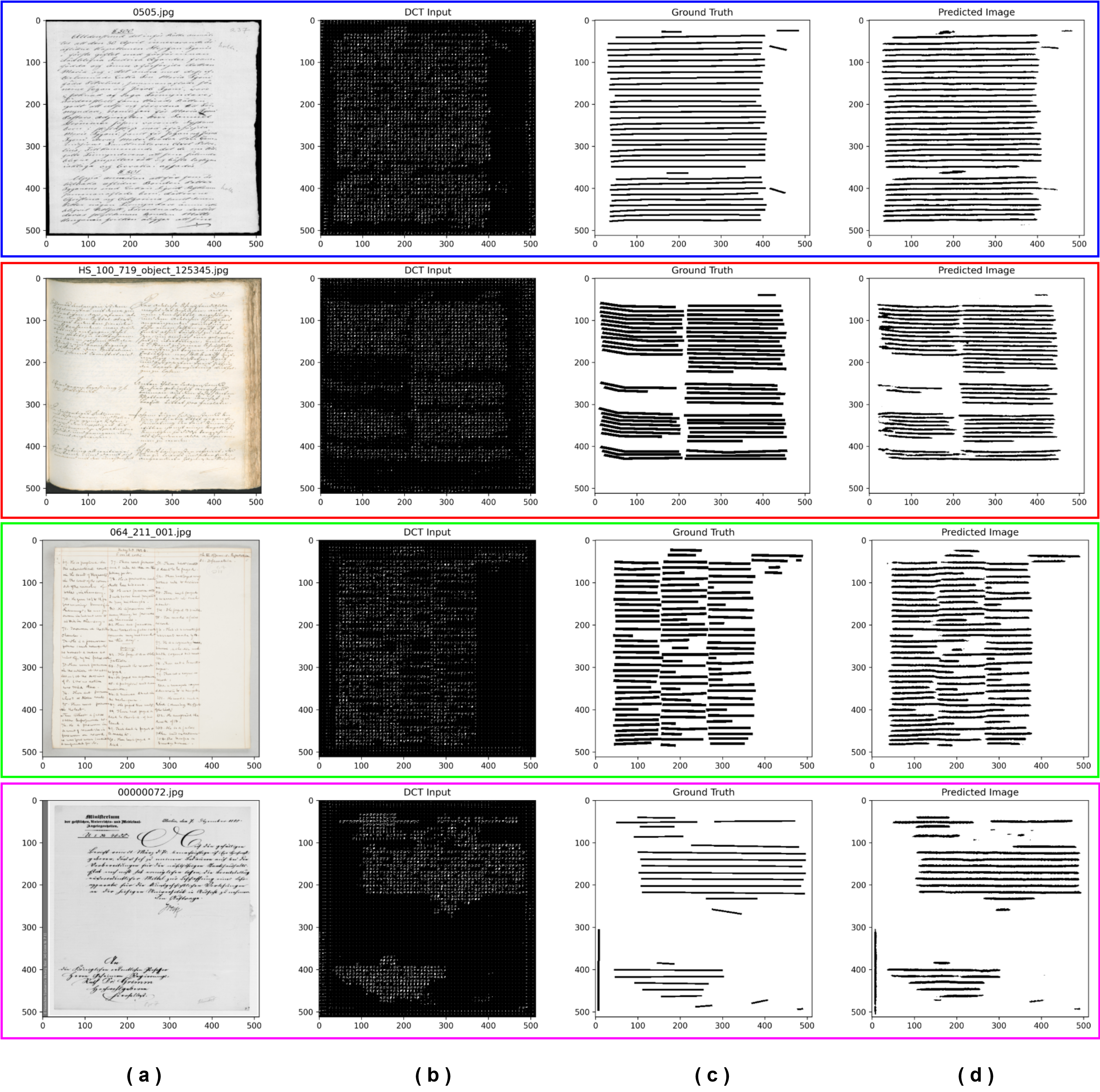}
    \caption{Some of the sample (a) uncompressed document images with their corresponding (b) JPEG compressed input versions, and (c) the actual ground truths of text-lines, (d) output documents with predicted text-lines locations. First two rows (blue and red colors) are outputs on simple documents and next two rows (green and rose colors) are outputs on complex documents}
    \label{figurefour}
\end{figure*}

\begin{figure*}[!ht]
    \centering
    \begin{tabular}{cc}
     \includegraphics[scale=0.10]{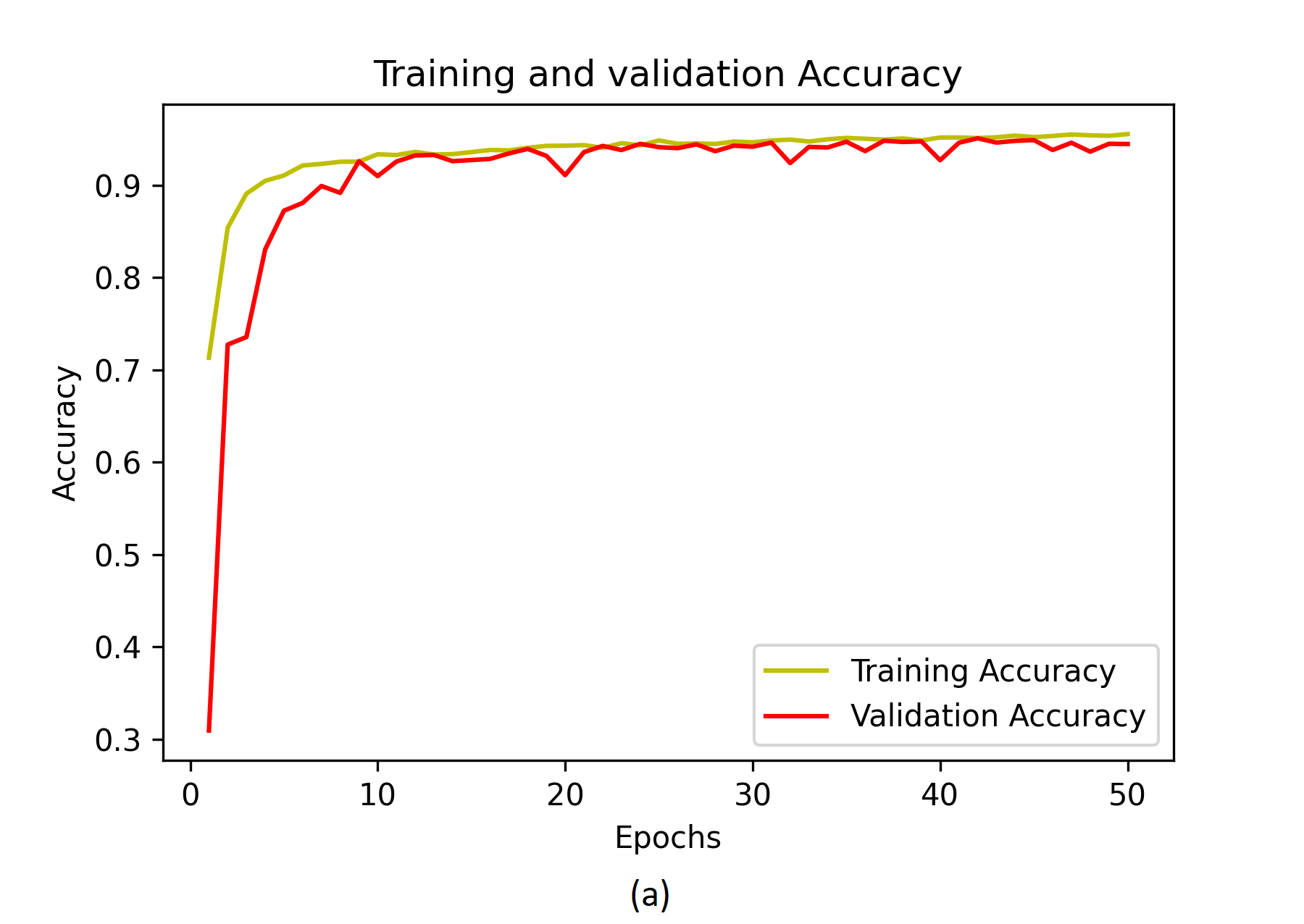}& \hspace{-10pt}\includegraphics[scale=0.10]{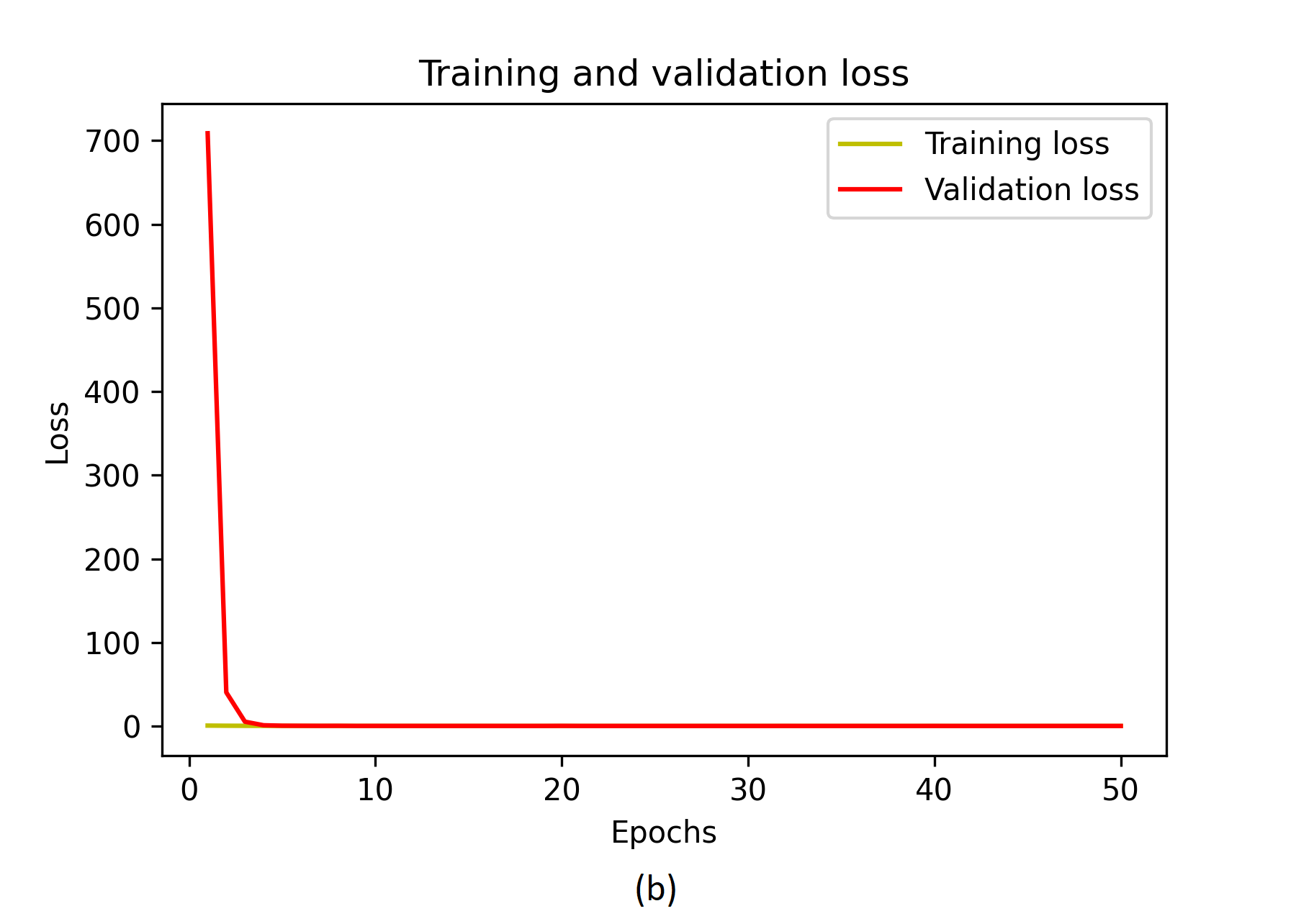}  \\
    \end{tabular}
    \caption{The (a) accuracy and (b) loss details of the proposed model tested on medium-resolution images (512$\times$512) for Track A (simple documents) images of ICDAR2017}
    \label{figurefive}
\end{figure*}

\begin{figure*}[!ht]
    \centering
    \begin{tabular}{cc}
    \includegraphics[scale=0.10]{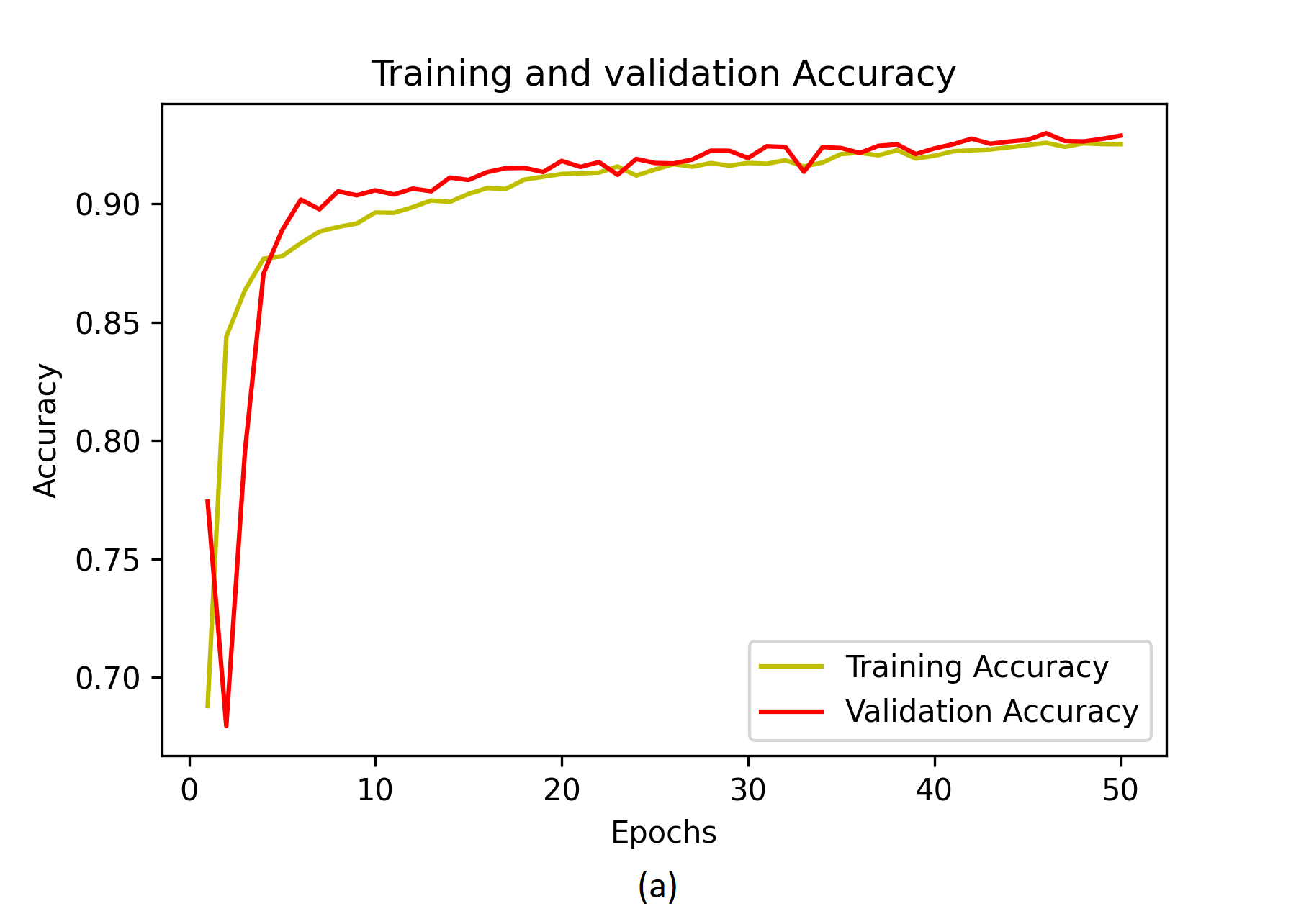}&
    \hspace{-10pt}\includegraphics[scale=0.10]{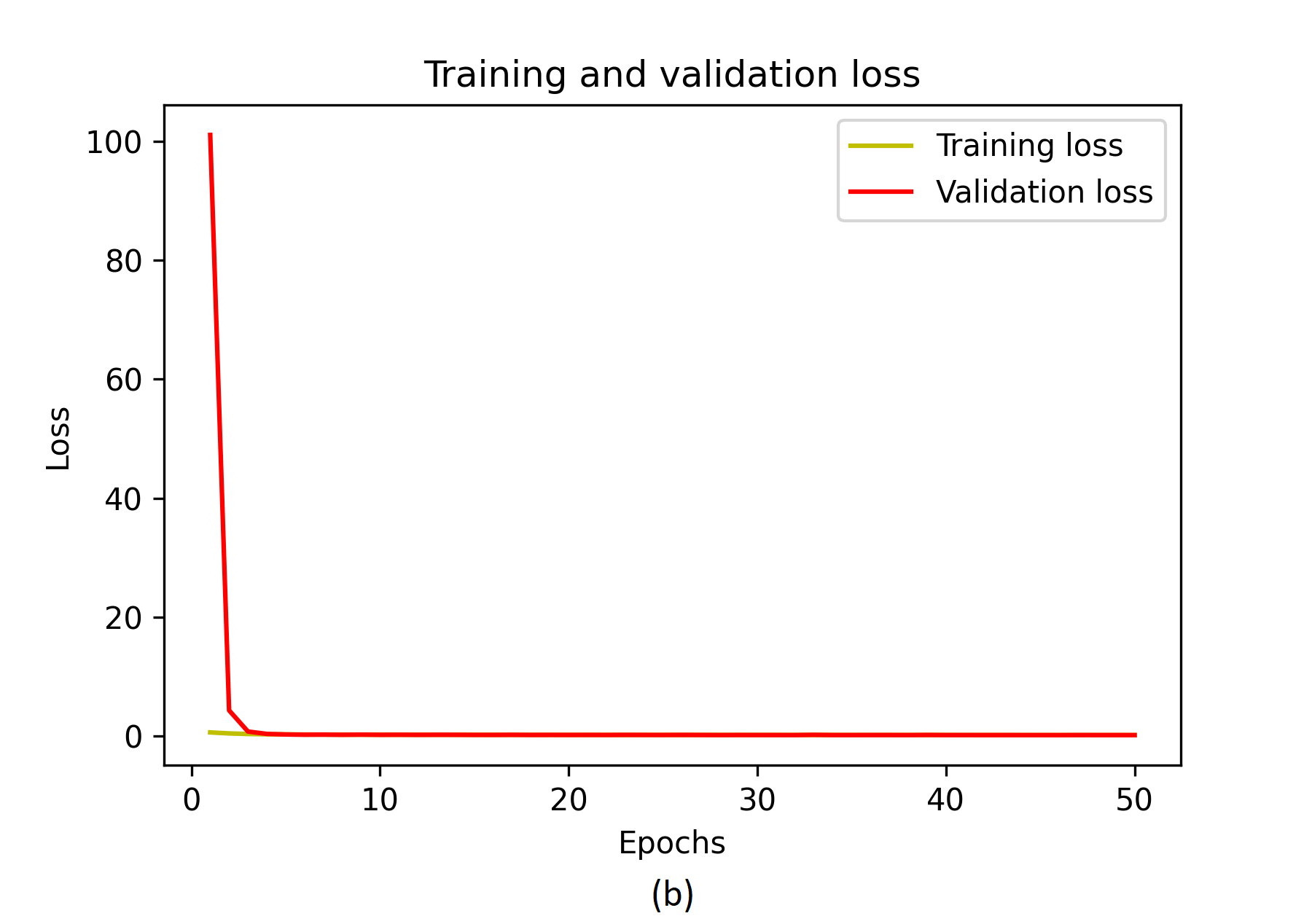}  \\
    \end{tabular}
        \caption{The (a) accuracy and (b) loss details of the proposed model tested on medium-resolution images (512$\times$512) for Track B (complex documents) images of ICDAR2017}
    \label{figuresix}
\end{figure*}

The experimental results of the proposed model tested on ICDAR2017 dataset are shown in Figure \ref{figurefour}, where \ref{figurefour}(a) is document image in uncompressed domain and \ref{figurefour}(b) same image with JPEG compressed input stream, \ref{figurefour}(c) is the ground truth along with its predicted output in \ref{figurefour}(d). It can be noticed that the proposed model has localized the text-line regions from both simple (first and second rows) and complex (third and fourth rows) documents, and in the midst of single, multi-columned layouts directly in JPEG compressed domain as shown in Figure \ref{figurefour}. The overall results of model tested on two datasets and evaluated by standard metrics are shown in Table \ref{tabletwo}. The model has achieved 96.4\% dice score and 93.4\% IoU on complex images of ICDAR2017 dataset and 96.4\% dice score and 93.2\% IoU on ICDAR2019 dataset. The overall performance of the proposed model and loss against each epoch are shown in Figure \ref{figurefive} and Figure \ref{figuresix} on both Track A and Track B sets. Since input image had some noise, there we performed post processing to improve the dice and IoU scores.

\begin{table*}[!ht]
\centering

\caption{Evaluation of the proposed model on CompTLL-UNet ICDAR-cBAD DCT domain}
\resizebox{\linewidth}{!}{
\begin{tabular}{ccccccc} 
 \hline
 Input&\textbf{\small{Track}} & \textbf{\small{Precision (\%)}} & \textbf{\small{Recall   (\%)}} & \textbf{\small{F-Measure (\%)}} & \textbf{\small{Dice Score (\%)}} & \textbf{\small{IoU (\%)}}   \\ 
 \hline
\small{ICDAR2017}&Simple layout & 96.0 & 97.0 & 96.0  & 97.2 & 94.0\\
\cline{2-7} 
{\textbf{\small{}}}&Complex layout & 95.0 & 97.0 & 96.0 & 96.4 & 93.4 \\
 \hline
\small{ICDAR2019}&Combined & 95.0 & 96.0 & 96.0  & 96.4 & 93.2\\
 \hline
\end{tabular}}
\label{tabletwo}
\end{table*}

The performance of the proposed model has been compared to different state-of-the-art methods existed both in the pixel domain and compressed domain as tabulated in Table \ref{tablethree}. In the table the methods in first 6 rows are in pixel domain and next two rows are in compressed domain. Most of these methods have been tested on simple and less constrained document images. Similarly the performance of the proposed model in terms of DICE score and IoU has been compared with the state-of-the-art result in Table \ref{tableforIOUanddice}. If we observe the results in the Table \ref{tablethree} and in Table \ref{tableforIOUanddice}, it is noticed that the results of the proposed model has outperformed both traditional and deep learning methods both in pixel domain and compressed domain, and showed state-of-the-art performance on challenging handwritten document images in JPEG compressed domain. 

\begin{table*}[!ht]
\centering
\caption{Comparison of the performance of the proposed model with existing methods tested on different handwritten document images in both pixel and compressed domains.}
\resizebox{\linewidth}{!}{
\begin{tabular}{ccccccc} 
 \hline
 \textbf{\small{Algorithm}} & \textbf{\small{Domain}}&
 \textbf{\small{Dataset}}& \textbf{\small{Model}} & \textbf{\small{Precision}} & \textbf{\small{Recall}} & \textbf{\small{F-Measure}}  \\
  & && \textbf{\small{Type}} &(\%) &(\%) &(\%)  \\
    \hline
    Kiumarsi et al. \cite{8583768} &Pixel&ICDAR2013&CC& 96.37 & 96.26 & 96.32\\
       \hline
    Renton et al \cite{renton2018fully}&Pixel&cBAD&FCN&94.9&88.1&91.3\\
    \hline
    Barakat et al. \cite{text-linefully} &Pixel&IHP&FCN& 82.0 & 78.0 & 80.0\\
     \hline
    Mechi et al. \cite{8978147}&Pixel&cBAD&A-UNet&75.0&85.0&79.0\\
    \hline
    Gader et al. \cite{9257759}&Pixel&BADAM&AR2UNet&93.2&94.3&93.7\\
     \hline
    Demir et al. \cite{9548523}&Pixel&IHP&GAN&83.0&88.0&85.0\\
    \hline
    Amarnath et al \cite{amarnath2018text} &Compressed&ICDAR2013&Handcrafted&95.8&89.2&92.4\\
    &(Run length)&&&&&\\
     \hline
    Rajesh et al \cite{bulla2020}&Compressed&ICDAR2013&Handcrafted&98.40&96.7&97.5\\
    &(JPEG)&&&&&\\
    \hline
    \textbf{\small{Proposed Method}}&Compressed&cBAD&\textbf{\small{CompTLL-UNet}}& 95.5 & 97.0 & 96.0 \\
    &\textbf{\small{(JPEG)}}&(ICDAR2017)&(Deeplearning)&&&\\
    \hline
    \textbf{\small{Proposed Method}}&Compressed&cBAD&\textbf{\small{CompTLL-UNet}}& 95.0 & 96.0 & 96.0 \\
    &\textbf{\small{(JPEG)}}&(ICDAR2019)&(Deeplearning)&&&\\
    \hline
\end{tabular}
}
 \label{tablethree}
\end{table*}

\begin{table*}[!ht]
\centering
\caption{Comparing the performance of proposed method with existing state-of-the-art methods in terms of DICE score and IoU.}
\begin{tabular}{ccccc} 
 \hline
 \textbf{\small{Model}} &\textbf{\small{Input Type}}& \textbf{\small{Dataset}} & \textbf{\small{DICE Score}} & \textbf{\small{IoU}}  \\ 
 \hline
 Mechi et al. \cite{8978147}& Pixel domain&cBAD &-& 65.0 \\ 

 \hline
 \textbf{\small{Proposed Method}}&Compressed&cBAD&96.4 & 93.4 \\
   &&(ICDAR2017)&& \\ 
 \hline
\textbf{\small{Proposed Method}}&Compressed&cBAD & 96.4 & 93.2\\
  &&(ICDAR2019)&& \\ 
   \hline
\end{tabular}
\label{tableforIOUanddice}
\end{table*}

\begin{figure*}[!ht]
    \centering
    \includegraphics[scale=0.35]{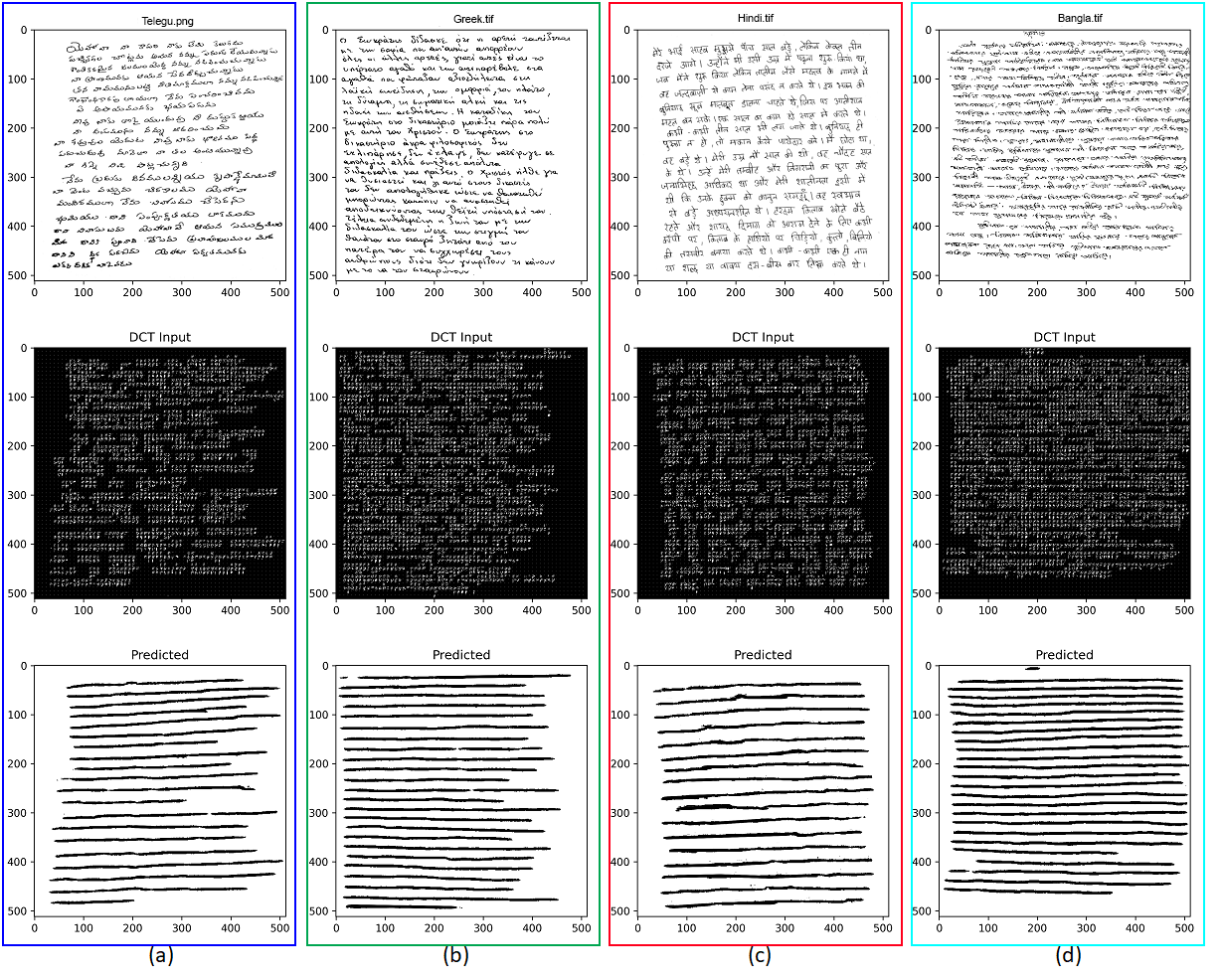}
    \caption{Segmentation results on various document images containing text-lines in various scripts such as (a) Telugu, (b) Greek, (c) Hindi, (d) Bangla in each column respectively. }
    \label{figseven}
\end{figure*}

In continuation to above experiments, the proposed model has been tested on document images with different scripts such as Telugu, Greek, Hindi, Bangla. The output images are shown in Figure \ref{figseven}. In the figure, the first column displays text-line in input document with Telugu script, and same document in JPEG compressed representation, and predicted output documents with segmented text-lines. Similarly second column contain document with Greek text-lines. Third row contain document with Hindi text-lines and fourth contain Bangla text-lines along with their predicted results. Based on these results, it can be understood that the proposed model can be applicable to segment text-lines in document images with different scripts directly in the compressed domain. Since the proposed model has been fed with direct DCT coefficients, we have conducted experiments by analysing the JPEG algorithm manually \cite{JPEGalgo} and noted the computational and storage gain achieved by feeding the JPEG compressed streams. We have observed the computational gain the model achieved based on a batch of 20 images for one complete epoch. The model has gained 20.2\% of computational gain in comparison to uncompressed domain. And since the model is fed with direct compressed data we reduce the decompression cost by 73.27\% , as shown in the Table \ref{tablefour}. Similarly, we have also calculated storage costs with respect to compressed domain by feeding the batch of images and noted the GPU storage costs. In this case the model has shown significant performance with 97.1\% reduction in storage. All the details of these experiments are tabulated in Table \ref{tablefour}.

\begin{table}[!ht]
\centering
\caption{Evaluation of the proposed model in terms of computational and storage costs with respect to feeding uncompressed document images versus JPEG compressed document images (20 images as batch size of 5) }
\begin{tabular}{cccc} 
 \hline

 \small{Complexity} &  \small{Pixel} & \small{DCT} & \% of  \\ 
 \small{Type} &  \small{ Domain } & \small{Domain} & Reduction  \\ 
 \hline
Computational & 7.27 sec & 5.8 sec & 20.2\% \\
 \hline
 Decompression cost& 107 sec & 28.6 sec & 73.27\%\\
 \hline
Storage & 22.5MB & 0.54 & 97.1\% \\
 \hline

\end{tabular}
\label{tablefour}
\end{table}

\begin{table*}[!htb]
\centering%
\caption{The experimental results of the proposed model tested on high-resolution (1024 $\times$ 1024), medium-resolution (512 $\times$ 512) and low-resolution (256 $\times$ 256) images extracted from the ICDAR2017 and ICDAR2019 datasets.}
\begin{tabular}{cccccccc}
\hline \\
Dataset &Layout & Input size& Precision & Recall& F-Measure & Dice Score& IoU \\
\hline
ICDAR2017 &Simple & 256$\times$256 & 95 & 96& 95 & 95.8& 93.1 \\
                 &Layout& 512$\times$512 & 96 & 97& 96 & 97.2& 94 \\
                 && 1024$\times$1024 & 97& 97 & 97& 97.8 & 94.5 \\
 &Complex & 256$\times$256 & 95 & 95& 95 & 95.1& 92.6 \\
                 &Layout& 512$\times$512 & 95& 97 & 96& 96.4 & 93.4 \\
                 && 1024$\times$1024 & 96& 97 & 96& 97.1 & 94.2 \\
ICDAR2019 &All& 256$\times$256 & 95 & 96& 96 & 94.7 & 92.8\\
                 &Layouts& 512$\times$512 & 95 & 96& 96 & 96.4& 93.2 \\
                 & & 1024$\times$1024 & 96& 97& 97& 97& 94 \\
\hline
\end{tabular}
\label{highandlow}
\end{table*}


\begin{figure*}[!ht]
    \centering
    \includegraphics[scale=0.21]{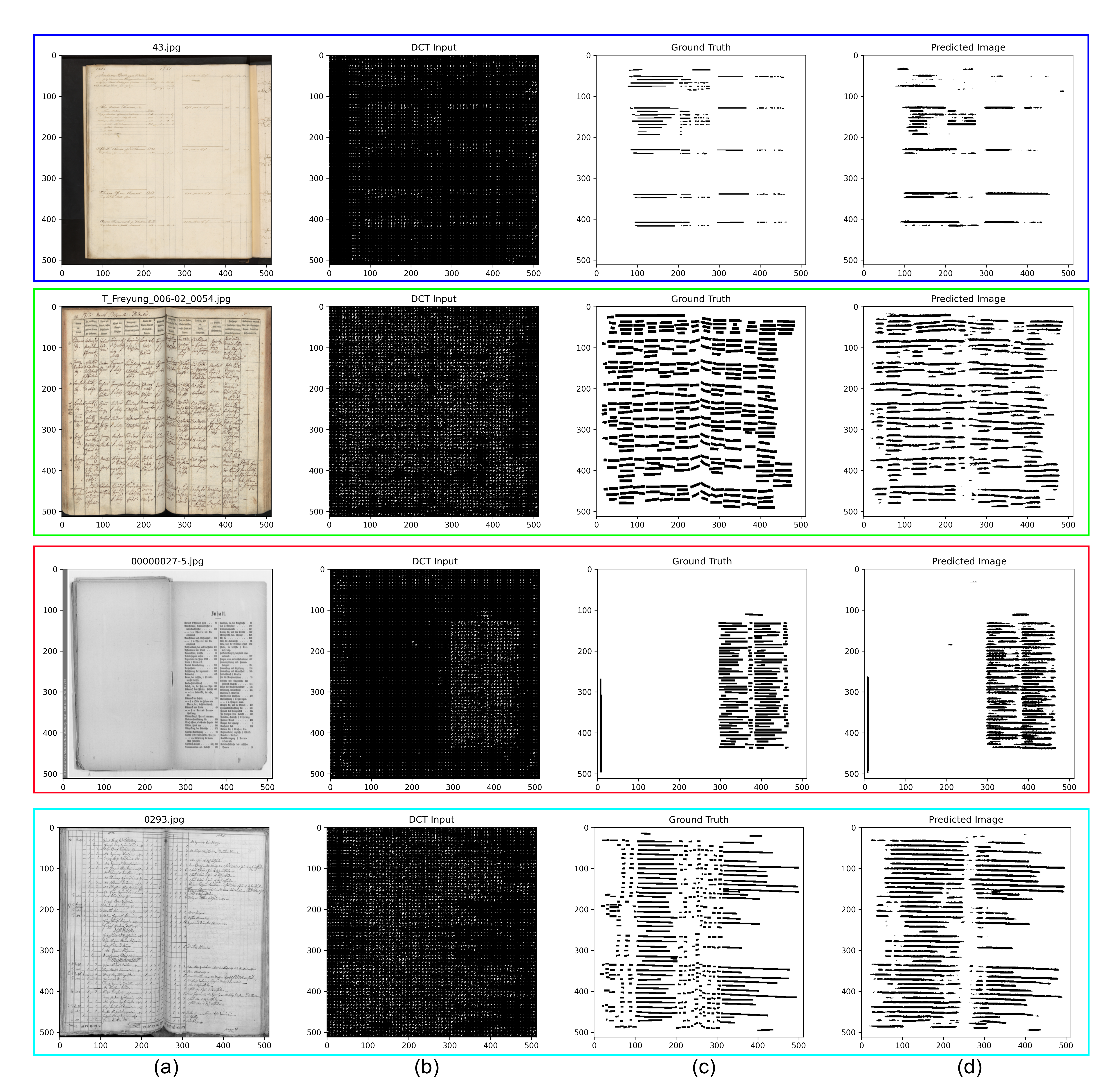}
    \caption{The experimental results of the proposed model tested on few of the (a) more challenging input document images, where (b) is compressed document images, and (c) is ground truth images and (d) is predicted output images in the compressed domain. }
    \label{error_cases}
\end{figure*}

In a practical scenario, oftentimes when the same image is transferred to various devices that have different display sizes. Due to such changes, the image resolutions shall be altered, and the original contents in the input image get modified. The contents within the low-resolution images are not as clear as in high-resolution images and may lose many important details when compared to high-resolution images. In order to overcome such practical challenges, the proposed deep learning model CompTLL-UNet has been analyzed with compressed input document images at different resolutions such as high-resolution ($1024\times1024$), medium-resolution ($512\times512$), and low-resolution ($250\times250$). The experimental results on the medium-resolution ($512 \times 512 $) are shown in the above section. In order to perform these experiments, we have converted the document images in ICDAR2017 and ICDAR2019 datasets into ($1024 \times 1024$) and ($512\times512$) resolutions. The experimental results on the compressed version of these high and low-resolution images are tabulated in Table \ref{highandlow}. 
From Table \ref{highandlow}, it is noticed that the model has achieved the 97\% accuracy when the input resolution is increased, and the model has achieved 95\% accuracy when the input resolution is decreased. One point to be noted here is that though the details of the contents in the high-resolution images are greater, the time for training the proposed model with such images is very higher. 

\hspace{20pt}Apart from the above results, there are some more challenging cases where the proposed model is unable to locate the text-line regions in the compressed domain. Some of the cases are shown in Figure \ref{error_cases}. The corresponding compressed input images are shown in Figure \ref{error_cases} (a). It can be noticed that the text-lines in the predicted output images are not as clear as in the ground truth images, as shown in Figure \ref{error_cases} (c) and Figure \ref{error_cases} (d). There are two reasons for this, and one is because the number of samples of such complex document images are less due to that it is unable to learn properly. The second reason is that because of the 8$\times$8 block compressed representation, the adjacent text-lines get overlapped or mixed up into the same DCT block or the text-lines may be overlapped with other additional contents such as vertical/horizontal stripes and background noise as shown in Figure \ref{error_cases} (a). Because of these reasons, the model failed to get the details where contents are completely overlapped as shown in Figure \ref{error_cases} (b). The performance of the model can be increased by adding more number of samples for challenging documents along with some preprocessing to improve the quality of contents in the input documents.

\hspace{20pt}Overall, based on the above results, it can be understood that the proposed CompTLL-UNet model has achieved promising performance in localizing the text-line segments directly in the compressed domain. It has also explained the practical advantages of feeding the compressed data that reduce the computational costs by 20.2\% and storage by 89.1\%. We also anticipate that the proposed compressed domain model  for text-line segmentation is a better method in comparison to the pixel domain methods, and advantageous in terms of storage and computational costs for real-time applications.

\section{Conclusion}
This paper has discussed a technique for the localization of text-line regions in challenging handwritten document images directly in JPEG domain. The JPEG compressed stream is analyzed through proposed deep learning model CompTLL-UNet to localize the text-line regions. The model is tested on document images with different scripts such as Telugu, Greek, Hindi, and Bengali, and achieved a good performance. We have tested the computational and storage performance by feeding the compressed streams into the proposed model, and it is observed that the model could reduce the computational and storage costs by 20.02\% and 97.1\%. We anticipate that the proposed method might be helpful in avoiding computational expenses in real-time applications. 
 \bibliographystyle{unsrt}
 \bibliography{bulla}

\end{document}